\documentclass{article}
\usepackage{ijcai22}
\usepackage{amssymb} 
\usepackage{times}
\usepackage{soul}
\usepackage{url}
\usepackage[hidelinks]{hyperref}
\usepackage[utf8]{inputenc}
\usepackage[small]{caption}
\usepackage{graphicx}
\usepackage{amsmath}
\usepackage{amsthm}
\usepackage{booktabs}
\usepackage{algorithm}
\usepackage{algorithmic}
\usepackage{color}
\usepackage{amsmath}
\usepackage{bbm}

\title{Data Augmentation For Label Enhancement}

\author{
Zhiqiang Kou$^{1,2}$
\and
 Yuheng Jia$^{1,2}$\thanks{*Corresponding author}\and
 Jing Wang$^{1,2}$\and
 Boyu Shi $^{1,2}$\and
Xin Geng$^{1,2}$\thanks{*Corresponding author}
\affiliations
$^1$MOE Key Laboratory of Computer Network and Information Integration, China\\
$^2$ School of Computer Science and Engineering, Southeast University, Nanjing 210096, China
\emails
\{zhiqiang \_Kou, yhjia,  wangjing91, shiboyu, xgeng \}@seu.edu.com,
}

\begin{document}

\maketitle

\begin{abstract}
Label distribution  (LD) uses the description degree to describe instances, which provides more fine-grained supervision information when learning with label ambiguity. Nevertheless, LD is unavailable in many real-world applications. To obtain LD, label enhancement (LE) has emerged to recover LD from logical label. Existing LE approach have the following problems: (\textbf{i}) They use logical label to train mappings to LD, but the supervision information is too loose, which can lead to inaccurate model prediction;  (\textbf{ii})  They ignore feature redundancy and use the collected features directly. To solve (\textbf{i}), we use the topology of the feature space to generate more accurate label-confidence. To solve  (\textbf{ii}), we proposed a novel supervised LE dimensionality reduction approach, which projects the original data into a lower dimensional feature space. Combining the above two, we obtain the augmented data for LE. Further, we proposed a novel nonlinear LE model based on the label-confidence and reduced features. Extensive experiments on 12 real-world datasets are conducted and the results show that our method consistently outperforms the other five comparing approaches.
\end{abstract}

\section{Introduction}

\begin{figure}[t]
\centering
\includegraphics[width=0.3\textwidth]{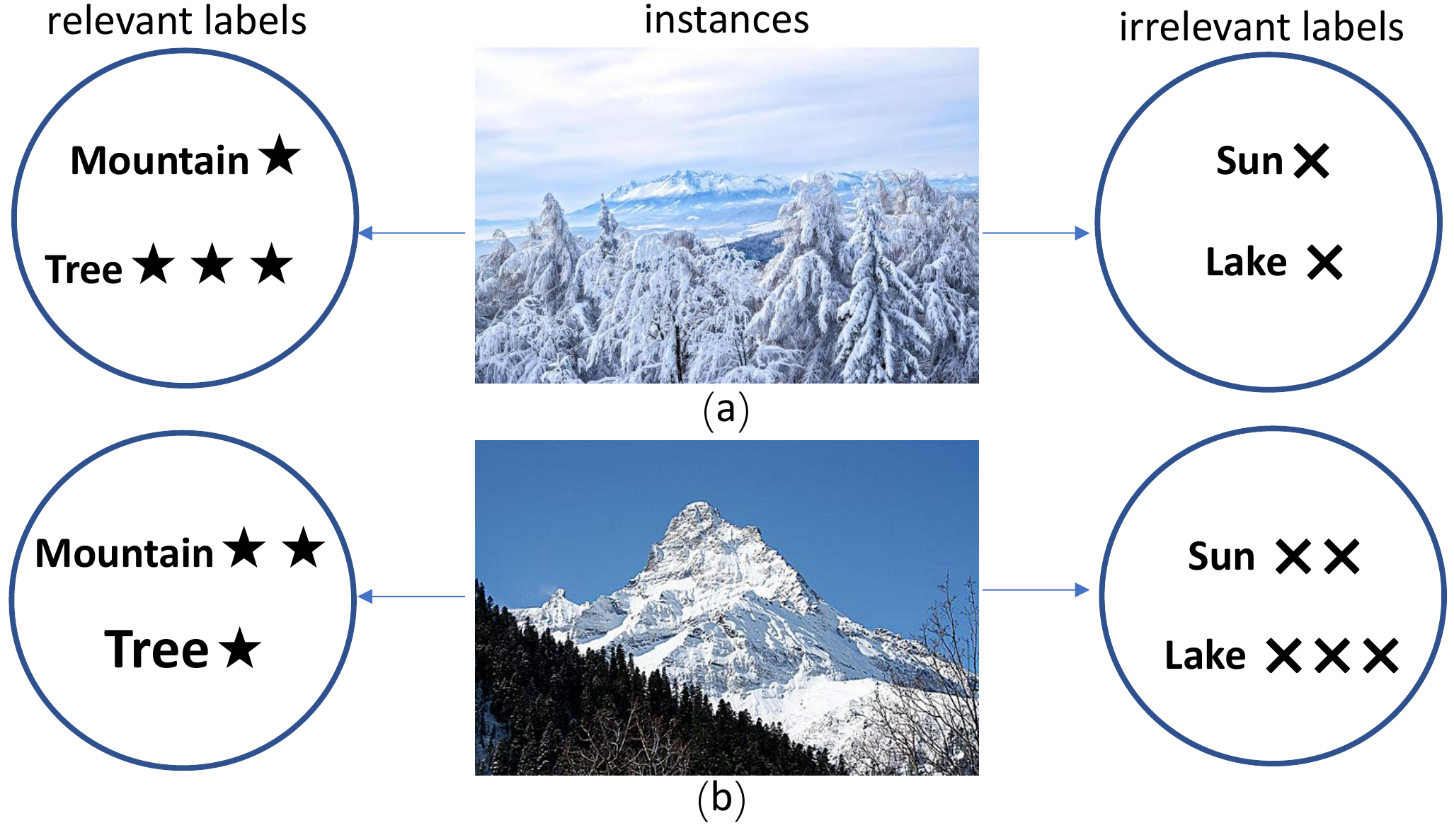} 
\caption{An example of the relative importance between relevant (irrelevant) labels between different instances}
\label{fig_ambig}
\end{figure}

 The process of learning is to build a mapping from instances to labels. The learning process of one sample corresponding to one label is not adequate for many tasks in the real world, based on which ambiguous learning is proposed, i.e. one instance corresponding to multiple class labels. The above learning process, also commonly referred to as multi-label learning (MLL). MLL uses strict logical label to divide the label into relevant and irrelevant label, however, the relevance or irrelevance of labels to instances is not absolute in real-world tasks. When multiple labels are associated with an instance, the relative importance between them may be different\cite{Xuicml}. As an example, when "mountain" and "tree" are relevant to the two images in Fig. \ref{fig_ambig}, it is clear that they differ in importance (in image(a), "tree" are more significant than "mountain", while in image(b) the opposite is true). At the same time, the importance of irrelevant labels may also vary, as shown in Fig. \ref{fig_ambig}, where "sun" and "lake" in image(a) are less irrelevant than in image(b). Therefore, directly labeling each relevant and irrelevant label using a logical label would ignore the relative importance of the label. 

To solve this problem,  Geng proposed a new learning paradigm called label distribution learning (LDL) \cite{geng2016label}. Different from MLL, LDL assumes that an instance
is annotated by an LD over all possible labels instead of a binary label vector. For example, give an instance $\mathbf{x}$, \textbf{$d_{\mathbf{x}}^y \in[0,1]$} denotes the description degree of the label $y$ to the instance $\mathbf{x}$, and $\sum_y d_{\mathbf{x}}^y=1$. LDL allows a better representation of the relativity between labels,  thus it has many real-world applications, such as pre-release prediction of crowd opinion on movies \cite{Geng2015PrereleasePO}, video parsing \cite{Geng2017SoftVP}, historical context-based style classification of painting images \cite{Yang2018HistoricalCS}, and children's empathy ability analysis \cite{Chen2021TowardCE}. However, LDs usually require extensive expert annotation, and therefore collecting LDs is very expensive. This led to the creation of an algorithm to recover LDs from training data in bulk, the process called label enhancement\cite{xu2019label}. In recent years, LE has been successfully applied to multi-label learning\cite{ijcai2022p524}, label distribution learning\cite{ijcai2020p446}, partial label learning\cite{DBLP:conf/aaai/XuLG19}, etc. 

\textbf{Motivation} Existing LE algorithms used the logical label directly to train a mapping to the LD and then normalize the predicted distribution. In classification tasks, logical label can provide accurate supervisory information, but when the goal is to recover an LD, it provides poor information.  In addition, the collected features are usually containing redundant information, the existing LE algorithm directly uses the collected features to train the model, which will lead to poor generalization of the model.

To deal with these challenges, we proposed a data augmented LE approach, LCdr.  Specifically, logical label provide little information, we utilised the  local consistency  of the instances  to learn a label-confidence that can provide more information. Next, for the first time  we proposed a supervised  dimensionality reduction approach in LE, which attemptes to projects the high-dimensional feature matrix containing redundant information into a low-dimensional space.  We called this process above  data-augmentation (DA). Finally, we proposed a non-linear LE model based on augmented data.  Extensive experiments on real-world   datasets demonstrate the superiority of the proposed approach and the effectiveness of augmented data. \\
\textbf{Our contribution is summarised below}:
\begin{itemize}
    \item For the first time, we use the label-confidence generated by the local consistency of instances to guide learning instead of logical label. This accurate supervisory information seems to improve the performance in LE.
    \item For the first time, we used  the dimensionality reduction technology in label enhancement,  and proved that DR can improve the generalizability of LE model.
    \item  Our approach achieves superior performance against five
existing LE algorithms on  twelve real-world datasets, and the proposed approach seems to   cooperate   with any LE algorithm.

\end{itemize}
\section{Related Work}
\subsection{Label Distribution Learning}
Label Distribution Learning (LDL) \cite{geng2016label} as a new learning paradigm, using LD to labeling instances and learns a mappings from instances to LD. Thus, in learning with label ambiguity task, the LD describes the supervised 
information at  a finer granularity than the logical label.  LDL has been successfully applied to many real  applications, such  as estimation of the
age of the speaker \cite{Si2022TowardsSA}, predictions of the composition of Martian craters \cite{Morrison2018PredictingMM}, facial depression recognition \cite{Zhou2022FacialDR}, joint acne image grading \cite{Wu2019JointAI}, depression detection \cite{Melo2019DepressionDB},  
indoor crowd counting \cite{Ling2019IndoorCC}, and head pose estimation \cite{Geng2020HeadPE}.  
\subsection{Label Enhancement}
The acquisition of LD often requires expert annotation, so the collection of LD is difficult. To obtain LD more easily, an algorithm for recovering the LD from the logical label  is proposed, 
and this algorithm is named Label Enhancement\cite{xu2019label}. 
In this section, we briefly review the researches in LE.  Label propagation techniques for generating label distributions in LP \cite{DBLP:journals/tkde/ZhangZFLG21} by propagating label importance information. ML \cite{DBLP:conf/aaai/HouGZ16} generates label distributions through the idea of manifold learning by means of the technique of locality embedding. LEMLL \cite{DBLP:conf/icdm/ShaoXG18} assumed that the label distribution space should share a similar local topology in the feature space, and a locally linear embedding technique is used to implement LE.  GLLE \cite{Xu2018LabelEF} constructed a local similarity matrix to preserve the structure of the feature space, and use the local similarity matrix to transform logical labels into label distributions. LESC \cite{Zheng2020GeneralizedLE} proposed a low-rank representation of LE by capturing the global relationships of samples and predicting implicit label correlations. LEVI \cite{xupami}  used variational inference techniques applied to LE to introduce a generative model of the label distributions while giving a variational lower bound on the recovered label distributions.

Although a number of LE algorithms have been proposed, there are no LE algorithms designed from a data representation perspective. The label space can be thought of as a low-dimensional representation of the feature space, where the collected feature and label are recorded as a set of data. A better representation of the data will undoubtedly improve the performance for LE.  In the next section, a novel data augmentation (DA) approach is proposed for LE.

\section{Data Augmentation for Label Enhancement}
\textbf{Problem  Setup} 
The goal of LE is to recover a label distribution matrix $\mathbf{D}=\left[\mathbf{d}_{1}, \mathbf{d}_{2}, \ldots, \mathbf{d}_{n}\right] \in \mathbbm{R}^{q \times n}$ from the training set $\mathcal{T}=\mathbf{(X, L)}$. We denote $Y=\left\{y_{1}, y_{2}, \ldots, y_{q}\right\}$ as the label space with q class labels, $\mathbf{d}_{i}=\left[d_{\mathbf{x}_{i}}^{y_{1}}, d_{\mathbf{x}_{i}}^{y_{2}}, \ldots, d_{\mathbf{x}_{i}}^{y_{q}}\right]\in \mathbf{D}$ as the label distribution vector to the $i$th sample $\mathbf{x}_{i}$, and $d_{\mathbf{x}_{i}}^{y}$ indicates the importance degree of label $y$ to $\mathbf{x}_{i}$. We denote  $\mathbf{X}=\left[\mathbf{x}_{1}, \mathbf{x}_{2}, \ldots, \mathbf{x}_{n}\right] \in \mathbbm{R}^{d\times n}$ as feature matrix and $\mathbf{L}=\left[\mathbf{l}_{1}, \mathbf{l}_{2}, \ldots, \mathbf{l}_{n}\right] \in \mathbbm{R}^{q \times n}$    as the corresponding  logical label   matrix, where $\mathbf{{l}}_{i}=\left(l_{{\mathbf{x}}_i}^{y_1}, l_{{\mathbf{x}}_i}^{y_2}, \ldots, l_{{\mathbf{x}}_i}^{y_q}\right)^{\top}\in \mathbf{L}$  is the the logical label vector. We denote $n$ and $d$ as the number of training examples and the dimension of features.\\
\textbf{Main Idea} 
Logical label ensures that each instance is correctly classified with a large  \textcolor{magenta}{ margin}.  When the goal of learning is to predict the LD of an instance, we need a more compact label to supervise training.  In addition, LE using the collected features directly, without considering the issue of feature  \textcolor{magenta}{ redundancy}, can reduced generalization ability of algorithms.  To solve the above problem, first we use the topology of the feature space to learn a more compact label-confidence, and then we use the LC to supervise the dimensionality reduction of the feature space. To verify whether the augmented  data can learn a reasonable LE model, we proposed a  augmented data based non-linear  LE approach.
\subsection{Generating Lable-Confidence  Utilizing Local Consistency of Instances}
Given an  training data set $\mathcal{T}$, we aim to generate a normalized real-valued label-confidence (LC) matrix $\mathbf{F}=\left[\mathbf{f}_1, \mathbf{f}_2, \ldots, \mathbf{f}_n\right]^{\top} \in[0,1]^{n \times q}$. Where  $ {\mathbf{f}_i=\left[f_{i 1}, f_{i 2}, \ldots, f_{i q}\right]^{\top}} $ represents the LC vector, and $f_{il}$ represents the probability of  the $y_l$  being  the ground-turth label of $\mathbf{x}_i$.  The label-confidence vector $\mathbf{f}_i$ satisfies the following constraints: (i) $\mathbf{x}_i(1 \leq i \leq n)$ with $f_{i l} \in[0,1]$,  (ii) $\sum_{l=1}^q f_{i l}=1$. Correspondingly, The LC matrix is initialized as:
\begin{equation}
\begin{gathered}
\forall 1 \leq i \leq n .\left\{\begin{array}{lll}f_{i l}=1 & \text { if } & l_{i l}=1 \\ f_{i l}=0 & \text { if } & l_{i l}=0 .\end{array}\right.
\end{gathered}
\label{chushihua}
\end{equation}
Here, we use the local consistency of the feature space to estimate the LC. For multi-label learning, this method has been proven to be effective in estimating the underlying ground-truth label. Give a fully
connected graph  $\mathcal{G}=(\mathcal{V}, \mathcal{E}, \mathbf{W})$, the weight matrix $\mathbf{{W}}=\left[w_{i j}\right]_{n \times n}$ over the LE training example is defined as: $w_{i j}=\exp \left(-\left\|\mathbf{x}_i-\mathbf{x}_j\right\|_2^2 / \sigma^2\right)$ if $\mathbf{x}_j\in \mathcal{N}\left(\mathbf{x}_i\right)$, otherwise $w_{ij}=0$, which $\sigma$ is the hyper-parameter. To ensure that our weight structure is symmetrical, we set $\mathbf{W}=\mathbf{W}+\mathbf{W}^{\top}$.  After  obtaining the local weight structure, we can obtain the label-confidence matrix by solving the following problem:
\begin{equation}
\begin{gathered}
\min _\mathbf{F} \sum_{i=1}^n \sum_{j=1}^n w_{i j}\left\|\frac{\mathbf{f}_i}{\sqrt{d_{i i}}}-\frac{\mathbf{f}_j}{\sqrt{d_{j j}}}\right\|_2^2 \\
\text { s.t. }  \sum_{l=1}^q f_{i l}=1, \quad 1 \leq i \leq n\\
 f_{i l} \geq 0\left(\forall l_{i l}=1\right), f_{i l}=0\left(\forall l_{i l}=0\right) ,
\end{gathered}
\label{LCLOSS}
\end{equation}
where $d_{i i}=\sum_{j=1}^n w_{i j}$ is the  degre of $\mathbf{x}_i$ in the graph.  The first term in Eq .\ref{LCLOSS} enforces a degree of similarity between the label confidece of nearby points. The first constraint represents the normalized property of the label-confidence vector. The second constraint represents the label-confidence is non-negative if it logical label is 1, and the label-confidence is zero if it Logical label is 0.   The optimization problem in Eq. \ref{LCLOSS} turns out to be:
\begin{equation}
\begin{gathered}
\min _{\widetilde{\mathbf{f}}} \frac{1}{2} \widetilde{\mathbf{f}}^{\top} \mathbf{H} \widetilde{\mathbf{f}}\\
 \text { s.t. } \mathbf{F} \mathbf{1}_q=\mathbf{1}_n, \mathbf{0}_{q \times n} \leq \mathbf{F} \leq \mathbf{Y},
\end{gathered}
\label{slove the LC}
\end{equation}
where $\widetilde{\mathbf{f}}=\operatorname{vec}(\mathbf{F})$, $\operatorname{vec}(\cdot)$ is the vectorization operator and $\mathbf{Q}\in \mathbb{R}^{nq\times n q}$ is defined as follows:

\begin{equation}
\begin{gathered}
\mathbf{Q}=\left[\begin{array}{cccc}
\mathbf{T} & \mathbf{0}_{n \times n} & \cdots & \mathbf{0}_{n \times n} \\
\mathbf{0}_{n \times n} & \mathbf{T} & \cdots & \mathbf{0}_{n \times n} \\
\vdots & \vdots & \ddots & \vdots \\
\mathbf{0}_{n \times n} & \mathbf{0}_{n \times n} & \cdots & \mathbf{T}
\end{array}\right],
\end{gathered}
\label{qpfloss}
\end{equation}
where $\mathbf{T} \in \mathbb{R}^{n \times n}$ is a square matrix defined as $\mathbf{T}=4\left(\mathbf{I}_{q \times q}-\mathbf{J}^{-\frac{1}{2}} \mathbf{W} \mathbf{J}^{-\frac{1}{2}}\right)$, where $\mathbf{J}$ is a diagonal matrix with its diagonal element defined as $d_{i i}=\sum_{j=1}^n w_{i j}$. Eq. \ref{slove the LC}  can be solved efficiently by off-the-shelf QP tools.  

By solving problem \ref{LCLOSS}, we obtain a augmented LC matrix $\mathbf{F}$, and we converted the training samples from $\mathcal{T}$ to $\widetilde{T}$, where $\widetilde{T}=\left\{\left(x_1, \boldsymbol{f}_1\right), \ldots,\left(x_n, \boldsymbol{f}_n\right)\right\}$. Based on the LC matrix, we introduce a supervised dimensionality reduction (DR) approach in the next section.

\subsection{Supervised Dimensionality Reduction for LE}

To remove redundant features, we proposed a supervised dimensionality reduction approach with maximum dependency between the feature information and the label space. \textit{It is not feasible to use MLL supervised DR method directly for label enhancement, because even if two instances are different, their corresponding binary labels may be same, and direct use of MLL DR method may loss of useful feature information}. Base on this, we proposed a  \textcolor{magenta}{ LC-based } DR approach for LE.

Specifically,  we consider a linear DR matrix $\mathbf{P}$, which can projection of feature  matrix $\mathbf{X}$  into a lower dimensional feature space, i.e. $\mathbf{P}^{\top}\mathbf{X}=\widetilde{\mathbf{X}} \in \mathbb{R}^{d^{\prime}\times n}(d^{\prime} \ll d)$.  First, we introduce the kernel matrix of input space $\mathbf{X}$, which is defined by $\mathbf{K}=\left[k_{i j}\right]_{n \times n}$, where $k_{ij}$ is defined as: $k_{i j}=k\left(\mathbf{x}_i, \mathbf{x}_j\right)=\left\langle\mathbf{P}^{\top} \mathbf{x}_i, \mathbf{P}^{\top} \mathbf{x}_j\right\rangle$,  next, locate the kernel matrix of the label-confidence   as $\mathbf{\widetilde{F}}=[\widetilde{f}_{ij}]_{n \times n}$, where $\widetilde{f}_{ij}$ is defined as: $\widetilde{f}_{ij}=l\left(\mathbf{f}_i, \mathbf{f}_j\right) =\left\langle\mathbf{f}_i, \mathbf{f}_j\right\rangle$.

Then we attempt to maximize the dependence between the projected feature  and the label-confidence. Here we use Hilbert-Schmidt Independence Criterion  (HSIC) \cite{Gretton2005MeasuringSD} \cite{zhang2010multilabel} \cite{bao2021partial}, which is an effective dependency measurement method. HSIC is defined as:
\begin{equation}
\begin{gathered}
\text{HSIC}(\mathcal{F}, \mathcal{U}, \widetilde{T})=(n-1)^{-2} \operatorname{tr}(\mathbf{H K H \mathbf{\widetilde{F}}}),
\end{gathered}bi
\label{yilai}
\end{equation}
where $\mathcal{F}, \mathcal{U}$ is the reproducing kernel Hilbert space mapped from $\mathbf{X}$ and $\mathbf{F}$ respectively,  tr($\cdot$) is the trace norm and $\mathbf{H}=\mathbf{I}-\frac{1}{m} \mathbf{e} \mathbf{e}^{\top}$, where $\mathbf{e}$ is an all-one column vector. However, there is still redundant information in the projected low-dimensional features. To maximize the retention of useful features, We constrain the projection vectors to be linearly uncorrelated with each other, i.e. $\mathbf{p}_i^{\top} \mathbf{X} \mathbf{X}^{\top}\mathbf{p}_j =\delta_{i j}$, where $\delta_{i j}$=1 if i=j and $\delta_{i j}$=0 otherwise.  Finally, we set the bases of the projection to be orthonormal, which is denoted as $\mathbf{p}_i^{\top} \mathbf{p}_j=\delta_{i j}$.  Substituting $\mathbf{K}=\mathbf{X}^{\top} \mathbf{P}\mathbf{P}^{\top} \mathbf{X}$ and the joint constraint into Eq. \ref{yilai} and  dropping the normalization term, we obtain the objective function as follows:
\begin{equation}
\begin{aligned}
\centering
 & \max _{\mathbf{P}}  \operatorname{tr}\left(\mathbf{H X}^{\top} \mathbf{P} \mathbf{P}^{\top} \mathbf{X H \widetilde{F}}\right) \\ 
&\text { s.t. }\alpha \mathbf{p}_i^{\top} \mathbf{X}\mathbf{X}^{\top} \mathbf{p}_j+(1-\alpha) \mathbf{p}_i^{\top} \mathbf{p}_j=\delta_{i j},
\end{aligned}
\label{HSCI}
\end{equation}
where $\alpha$ is trade-off parameter. The solution of Eq. \ref{HSCI} can be obtained easily by the Lagrangian method. Specifically, we set the derivative of the Lagrangian function of Eq. \ref{HSCI} to 0, and then  $\mathbf{P}$ can be obtained by find the eigenvectors ($\mathbf{V}$) of $\mathbf{XH\widetilde{F}H}\mathbf{X}^{\top}$ and  $\alpha \mathbf{XX}^{\top}+(1-\alpha) \mathbf{I}$, here we use the Matlab function $\mathbf{V}=$eig($\cdot$,$\cdot$). Then, we sort the matrix of eigenvalues in descending order (  $\mathbf{P}^{\prime}$=sort($\mathbf{V}$, 'descend')), and  select  the largest $d^{\prime}$ eigenvalues as the DR matrix,  i.e. 
\begin{equation}
\begin{aligned}
     \mathbf{P}=\mathbf{P}^{\prime}(:,1:d^{\prime}).
\end{aligned}
\label{yuzhicanshu}
\end{equation}

we can control the dimensionality of the projected feature space by setting the different threshold parameter $d^{\prime}$. By solving for the problem \ref{HSCI}, we obtain a low-dimensional feature matrix $\mathbf{\widetilde{X}}$, and combined with the previous section, we obtain the $\mathbf{augmented }$ $\mathbf{data}$, i.e.   $\mathbbm{T}^{\prime}=\left\{\left(\mathbf{\widetilde{x}}_i, \mathbf{f}_i\right) \mid 1 \leqslant i \leqslant n\right\}$, where $\mathbf{\widetilde{x}}_i=\mathbf{P}^{\top}\mathbf{x}_i$. In the next section, based on the augmented data, we proposed a label-confidence dimensionality reduction (LCdr) approach to recover label distribution.

\subsection{Non-linear LE Methods based on Augmented Data}
To verify the effectiveness  of the data augmentation, we proposed a new LE model named LCdr, which is defined as:
\begin{equation}
\mathcal{L}_{le}=\sum_{i=1}^n\left\|\mathbf{f}_i-\left[\sigma\left(\mathbf{w}_i \widetilde{\mathbf{x}}_i\right)_{}\right]_{\text {Softmax }}\right\|_2^2+\sum_{i=1}^n \beta\|\mathbf{w}_i\|_2^2,
\label{LEmodel}
\end{equation}
where $\mathbf{w}_i \in \mathbf{W}^{q\times d^{\prime}}$ is weighting matrix,  $\beta$  is a trade-off parameter,  $\sigma(\cdot)$ is  the activation function,  here, we use the Relu-function, and Softmax$(\cdot)$ is the normalization function.  The Relu-function  ensures that  the mapping between low-dimensional features and LC is non-linear, which allows our model to learn more complex relationships between the data. The Softmax$(\cdot)$  fuction   ensured  the learned distribution is non-zero and normalised.  The Second  term   of loss fuction \ref{LEmodel}  prevents our model from overfitting.

\textbf{ Implementation Details} We use the stochastic gradient descent (SGD) algorithm in PyTorch to solve Eq. \ref{LEmodel}, and  We set $\beta=0.1$, the learning rate to 0.01, the momentum to 0.9 and the weight-decay to 5e-4. The learning process continues until convergence or reaches predefined maximum iterations.

The model iterates until it converges, and finally, we directly predict the LD of the training instances.

Augmented data can cooperate with many existing LE approach such as GLLE \cite{xu2019label}, LESC \cite{Tang2020LabelEW}, LEMLL \cite{DBLP:conf/icdm/ShaoXG18}, LP \cite{DBLP:journals/tkde/ZhangZFLG21}, ML2 \cite{DBLP:conf/aaai/HouGZ16}.  Algorithm \ref{algorithm1} summarizes the pseudo-code of the proposed approach.

\begin{algorithm}[t]\scriptsize
\caption{The pseudo-code of the proposed approach}\label{algorithm: The porposed approach  }
 \textbf{Input}: \\
$\mathcal{T}$:  LE training set $\left\{\left(\mathbf{x_i}, \mathbf{l_i}\right) \mid 1 \leq i \leq n\right\}$\\
$k$: the number of  neighbors  for weighted graph construction\\
$d^{\prime}$: the threshold parameter in Eq. \ref{yuzhicanshu}\\
$\alpha$: the trade-off parameter in Eq. \ref{HSCI} \\
$\mathcal{F}$: the proposed approach, which is defined as Eq. \ref{LEmodel};\\
$\mathcal{R}$: any label enhancement  algorithm\\ 
 \textbf{Output}: \\
$\mathbf{D}_1$: the predicted LD  by $\mathcal{F}$,\\
$\mathbf{D}_2$: the predicted LD  by  $\mathcal{R}$;\\
 \textbf{Process}: 
 \begin{algorithmic}[1]
 \STATE Set the fully connected graph  $\mathcal{G}=(\mathcal{V}, \mathcal{E}, \mathbf{W})$, and calculate the weight matrix $\mathbf{W}$; \\
 \STATE Initialize the $q$×$n$ label-confidence matrix $\mathbf{F}$ based on Eq. \ref{chushihua};
\REPEAT
 \STATE Update $\mathbf{F}$ by solving  Eq.\ref{LCLOSS};  
  \UNTIL{convergence}
 \RETURN $\mathbf{F}$;
 \REPEAT
  \STATE Calculate kernel matrix $\mathbf{\widetilde{F}}=\left\langle\mathbf{f}_i, \mathbf{f}_j\right\rangle$ and $\mathbf{H}=\mathbf{I}-\frac{1}{m} \boldsymbol{e} \boldsymbol{e}^{\top}$;
 \STATE Solve the  problem  \ref{HSCI}, and then get the projection matrix $\mathbf{P}$;  
  \UNTIL{convergence }
 \RETURN the lower-dimensional features space : $\widetilde{\mathbf{X}}=\mathbf{P}^{\top} \mathbf{X}$;
 \STATE Transformed the augmented  LE training set  $\mathbbm{T}^{\prime}=(\widetilde{\mathbf{X}}, \mathbf{F})$;
\RETURN $\mathbf{D}_1=\mathcal{F}(\mathbbm{T}^{\prime})$;
 \RETURN $\mathbf{D}_2= \mathcal{R}(\mathbbm{T}^{\prime})$;

  \
 \end{algorithmic}
 \label{algorithm1}
\end{algorithm}

\section{Experiments}
\textbf{Datasets and evaluation measure}: We validate the effectiveness of LCdr on 12 real data sets, with the specific details of the data shown in Table\ref{detailsofdatasets}. Where the first 10 real-world datasets are collected from the records of 10 biological experiments on the budding yeast genes\cite{Eisen1998ClusterAA}, the last datasets  Flickr-LDL and Twitter-LDL\cite{aaai/YangSS17} are  are two  two facial expression datasets.  More details of them can be found in \cite{geng2016label}. Where the logical labels in the datasets are obtained by LD degradation, here we use the same strategy as GLLE\cite{Xu2018LabelEF}, which can be roughly described as setting a threshold, where logical labels with a degree of description more than the threshold are 1 and the rest are 0.   Following Geng \shortcite{geng2016label}, seven metrics are  used to evaluate the ability to recover, i.e.,  Chebyshev$\downarrow$, Clark$\downarrow$, Kullback-Leibler(KL)$\downarrow$,  Canberra$\downarrow$, S$\phi$rensen$\downarrow$ and Cosine$\uparrow$, ↑ (resp. ↓) means the larger (resp. smaller), the better, and the seven  measures are summarized in Table \ref{detailsofdatasets}.  

\begin{table}[]\scriptsize
\centering
\begin{tabular}{cc}
\hline\hline
Measure          & Formula \\ \hline
Chebyshev $\downarrow$       &   $\operatorname{Dis}_1(\boldsymbol{d}, \hat{\boldsymbol{d}})=\max _j\left|d_j-\hat{d}_j\right|$       \\ \hline
Clark            &  $\operatorname{Dis}_2(\boldsymbol{d}, \hat{\boldsymbol{d}})=\sqrt{\sum_{j=1}^c \frac{\left(d_j-\hat{d}_j\right)^2}{\left(d_j+\hat{d}_j\right)^2}}$        \\ \hline
Canberra $\downarrow$        &   $\operatorname{Dis}_3(\boldsymbol{d}, \hat{\boldsymbol{d}})=\sum_{j=1}^c \frac{\left|d_j-\hat{d}_j\right|}{d_j+\hat{d}_j}$      \\ \hline
Kullback-Leibler$\downarrow$ &    $\operatorname{Dis}_4(\boldsymbol{d}, \hat{\boldsymbol{d}})=\sum_{j=1}^c d_j \ln \frac{d_j}{\hat{d}_j}$    \\ \hline
cosine   & $\operatorname{Sim}_1(\boldsymbol{d}, \hat{\boldsymbol{d}})=\frac{\sum_{j=1}^c d_j \hat{d}_j}{\sqrt{\sum_{j=1}^c d_j^2} \sqrt{\sum_{j=1}^c \hat{d}_j^2}}$     \\ \hline

intersection $\uparrow$    &   $\operatorname{Sim}_2(\boldsymbol{d}, \hat{\boldsymbol{d}})=\sum_{j=1}^c \min \left(d_j, \hat{d}_j\right)$      \\ \hline\hline
\end{tabular}
\caption{The distribution distance/similarity measures}
\label{PINGJIAZHIBIAO}
\end{table}

\begin{table}[t]\small
\centering
\scalebox{0.9}{
\begin{tabular}{lllll}
\hline\hline
ID & Dataset     & Examples & Features & Labels \\ \hline
1  & Yeast-dtt (dtt)  & 2465      & 24        & 4       \\
2  & Yeast-spo (spo)  & 2465      & 24        & 6       \\
3  & Yeast-spo5 (spo5) & 2465      & 24        & 3       \\
4  & Yeast-diau (diau) & 2465      & 24        & 7       \\
5  & Yeast-cold (cold) & 2465      & 24        & 4       \\
6  & Yeast-spoem(spoem) & 2465      & 24        & 2       \\
7  & Yeast-cdc (cdc)  & 2465      & 24        & 15      \\
8  & Yeast-alpha (alpha) & 2465      & 24        & 18      \\
9  & Yeast-elu (elu)  & 2465      & 24        & 14      \\
10 & Yeast-heat (heat)     & 2465       & 24      &  5       \\
11 & Flickr-LDL (Flickr)  & 11150    & 200       & 8      \\
12 & Twitter-LDL(Twitter) & 10045     & 200       & 8      \\ \hline\hline
\end{tabular}}
\caption{Details of the datasets.}
\label{detailsofdatasets}
\end{table}

\textbf{Implementation Details}:  First we verify whether the proposed method can recover better LD from logical label than other LE algorithms. Next, we verify whether extending the data augmentation strategy to other LE algorithms enables it to learn a reasonable LE model.   Finally, we checked to ensure that the recovered LD could be trained to produce a more reasonable LDL model.  

 \begin{figure}[t]
\centering
\includegraphics[width=0.495\textwidth]{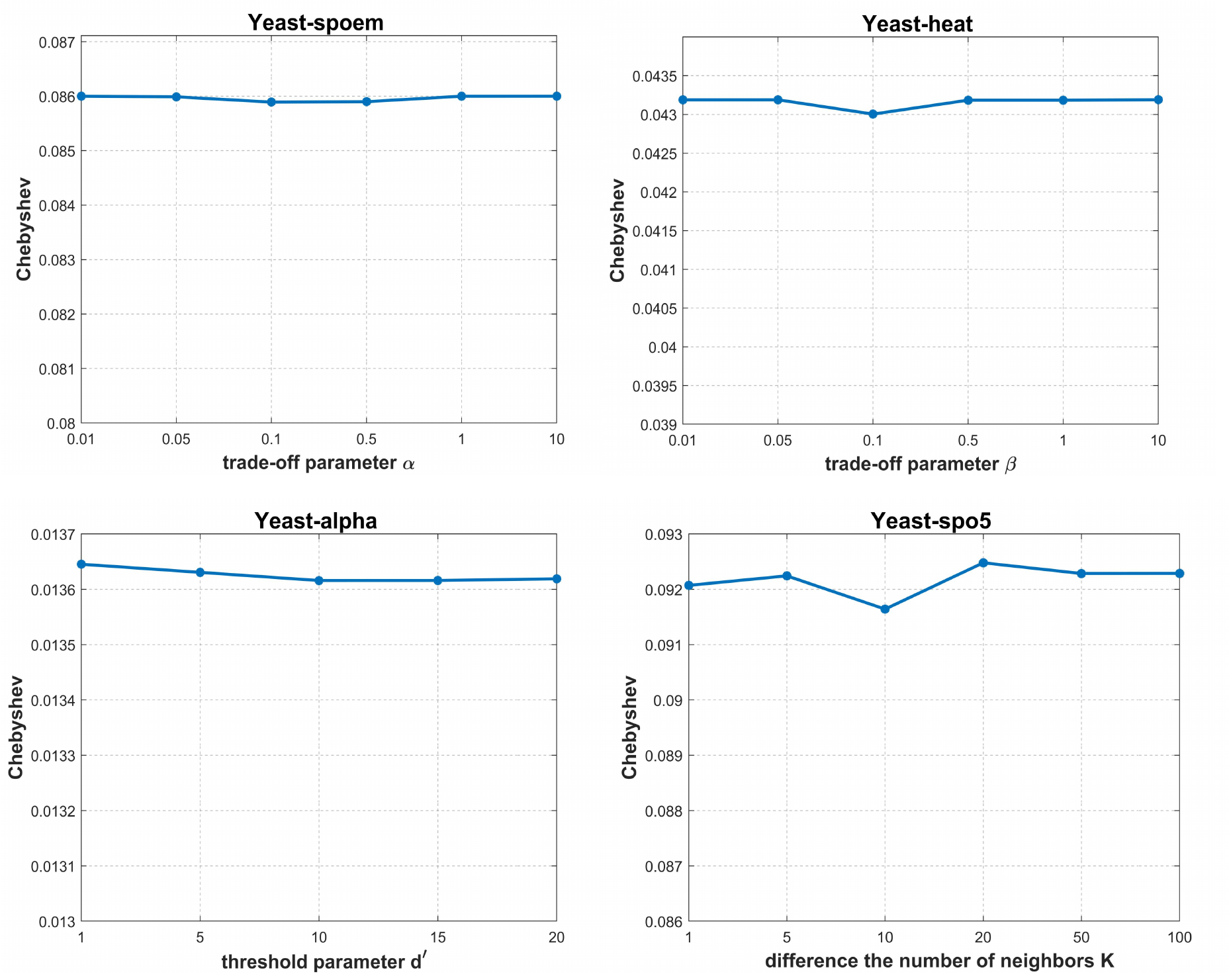} 
\caption{Performance of the proposed method as the  parameter $\alpha$, $\beta$,  $d^{\prime}$ and $K$ vary on different data sets.}
\label{para}
\end{figure}

\begin{table*}[t]\tiny\centering
\setlength{\tabcolsep}{3mm}
\begin{tabular}{lcccccccccccc}
\hline\hline
\multicolumn{13}{c}{chebyshev $\downarrow$} \\ \hline
& spoem           & alpha           & spo5            & cdc             & cold            & diau            & dtt             & elu             & heat            & spo             & flickr          & twitter         \\ \hline
ours      & \textbf{0.0859} & \textbf{0.0136} & \textbf{0.0916} & \textbf{0.0166} & \textbf{0.0535} & \textbf{0.0413} & \textbf{0.0366} & \textbf{0.0165} & \textbf{0.0432} & \textbf{0.0596} & \textbf{0.0614} & \textbf{0.0730} \\
LP        & 0.1286          & 0.0243          & 0.1040          & 0.0297          & 0.1002          & 0.0625          & 0.1017          & 0.0309          & 0.0699          & 0.0698          & 0.0661          & 0.0753          \\
ML2       & 0.3024          & 0.6264          & 0.2670          & 0.5394          & 0.2795          & 0.3823          & 0.2766          & 0.5208          & 0.3097          & 0.2835          & 0.1093          & 0.2467          \\
GLLE      & 0.0880          & 0.0190          & 0.0978          & 0.0216          & 0.0638          & 0.0515          & 0.0505          & 0.0223          & 0.0476          & 0.0599          & 0.0640          & 0.0782          \\
LESC      & 0.0892          & 0.0183          & 0.1013          & 0.0220          & 0.0631          & 0.0547          & 0.0520          & 0.0233          & 0.0504          & 0.0640          & 0.0837          & 0.0792          \\
LEMLL     & 0.1126          & 0.0397          & 0.1134          & 0.0442          & 0.1241          & 0.1045          & 0.1154          & 0.0479          & 0.0826          & 0.0777          & 0.0647          & 0.0815          \\ \hline
\multicolumn{13}{c}{clark $\downarrow$} \\ \hline
& spoem           & alpha           & spo5            & cdc             & cold            & diau            & dtt             & elu             & heat            & spo             & flickr          & twitter         \\ \hline
ours  & \textbf{0.1276} & \textbf{0.2148} & \textbf{0.1847} & \textbf{0.2190} & \textbf{0.1452} & \textbf{0.2206} & \textbf{0.0996} & \textbf{0.2042} & \textbf{0.1866} & \textbf{0.2536} & \textbf{0.2417} & \textbf{0.2649} \\
LP    & 0.2062          & 0.8226          & 0.2453          & 0.7369          & 0.3524          & 0.4669          & 0.3799          & 0.7119          & 0.4321          & 0.3971          & 0.4693          & 0.3881          \\
ML2   & 0.6195          & 3.4836          & 0.7556          & 2.9377          & 1.0435          & 1.6434          & 1.0491          & 2.7752          & 1.3994          & 1.3311          & 0.6439          & 1.4296          \\
GLLE  & 0.1313          & 0.3261          & 0.1942          & 0.3003          & 0.1704          & 0.2916          & 0.1380          & 0.2839          & 0.2064          & 0.2585          & 0.3058          & 0.3071          \\
LESC  & 0.1327          & 0.3085          & 0.2022          & 0.3020          & 0.1700          & 0.2934          & 0.1407          & 0.2938          & 0.2179          & 0.2744          & 0.4154          & 0.3174          \\
LEMLL & 0.1842          & 0.7693          & 0.2441          & 0.6702          & 0.3506          & 0.5848          & 0.3479          & 0.6604          & 0.3964          & 0.3724          & 0.3017          & 0.3058          \\ \hline
\multicolumn{13}{c}{canberra $\downarrow$} \\ \hline
 & spoem           & alpha           & spo5            & cdc             & cold            & diau            & dtt             & elu             & heat            & spo             & flickr          & twitter         \\ \hline
ours     & \textbf{0.1776} & \textbf{0.6984} & \textbf{0.2836} & \textbf{0.6545} & \textbf{0.2504} & \textbf{0.4730} & \textbf{0.1709} & \textbf{0.6010} & \textbf{0.3721} & \textbf{0.5233} & \textbf{0.5587} & \textbf{0.6135} \\
LP       & 0.2803          & 3.3451          & 0.3617          & 2.7223          & 0.6619          & 1.1169          & 0.7267          & 2.5437          & 0.9872          & 0.8897          & 1.2230          & 1.0005          \\
ML2      & 0.7892          & 14.1515         & 1.0781          & 10.7003         & 1.8913          & 3.9351          & 1.9135          & 9.7282          & 3.1178          & 2.9511          & 1.6257          & 3.8012          \\
GLLE     & 0.1825          & 1.0962          & 0.3012          & 0.9370          & 0.2957          & 0.6631          & 0.2401          & 0.8675          & 0.4170          & 0.5353          & 0.7208          & 0.7058          \\
LESC     & 0.1846          & 1.0455          & 0.3123          & 0.9352          & 0.2938          & 0.6443          & 0.2441          & 0.8908          & 0.4412          & 0.5653          & 0.9618          & 0.7219          \\
LEMLL    & 0.2491          & 2.8064          & 0.3688          & 2.2412          & 0.6195          & 1.3242          & 0.6225          & 2.1466          & 0.8450          & 0.7861          & 0.7360          & 0.7208          \\ \hline
\multicolumn{13}{c}{K-L  $\downarrow$} \\ \hline
& spoem           & alpha           & spo5            & cdc             & cold            & diau            & dtt             & elu             & heat            & spo             & flickr          & twitter         \\ \hline
ours   & \textbf{0.0274} & \textbf{0.0058} & \textbf{0.0317} & \textbf{0.0076} & \textbf{0.0135} & \textbf{0.0166} & \textbf{0.0066} & \textbf{0.0066} & \textbf{0.0137} & \textbf{0.0272} & \textbf{0.0182} & \textbf{0.0228} \\
LP     & 0.0449          & 0.0668          & 0.0363          & 0.0645          & 0.0566          & 0.0533          & 0.0651          & 0.0640          & 0.0558          & 0.0482          & 0.0530          & 0.0436          \\
ML2    & 0.2410          & 1.8433          & 0.2447          & 1.4508          & 0.3896          & 0.7066          & 0.4006          & 1.3624          & 0.5468          & 0.4789          & 0.0895          & 0.4202          \\
GLLE   & 0.0276          & 0.0123          & 0.0328          & 0.0130          & 0.0177          & 0.0259          & 0.0119          & 0.0124          & 0.0164          & 0.0274          & 0.0281          & 0.0301          \\
LESC   & 0.0284          & 0.0117          & 0.0364          & 0.0137          & 0.0180          & 0.0283          & 0.0128          & 0.0136          & 0.0184          & 0.0307          & 0.0523          & 0.0318          \\
LEMLL  & 0.0406          & 0.0629          & 0.0407          & 0.0582          & 0.0636          & 0.0900          & 0.0614          & 0.0610          & 0.0519          & 0.0466          & 0.0234          & 0.0281          \\ \hline
\multicolumn{13}{c}{cosine $\uparrow$} \\ \hline
& spoem           & alpha           & spo5            & cdc             & cold            & diau            & dtt             & elu             & heat            & spo             & flickr          & twitter         \\ \hline
ours   & \textbf{0.9783} & \textbf{0.9944} & \textbf{0.9738} & \textbf{0.9930} & \textbf{0.9875} & \textbf{0.9856} & \textbf{0.9940} & \textbf{0.9938} & \textbf{0.9873} & \textbf{0.9753} & \textbf{0.9778} & \textbf{0.9711} \\
LP     & 0.9664          & 0.9445          & 0.9740          & 0.9465          & 0.9546          & 0.9578          & 0.9470          & 0.9471          & 0.9547          & 0.9611          & 0.9554          & 0.9619          \\
ML2    & 0.8741          & 0.4002          & 0.8782 & 0.4791          & 0.8044          & 0.6917          & 0.7916          & 0.4993          & 0.7460          & 0.7773          & 0.9304          & 0.7982          \\
GLLE   & 0.9774          & 0.9880          & 0.9713          & 0.9877          & 0.9832          & 0.9760          & 0.9888          & 0.9879          & 0.9847          & 0.9753          & 0.9723          & 0.9634          \\
LESC   & 0.9769          & 0.9886          & 0.9687          & 0.9870          & 0.9831          & 0.9740          & 0.9879          & 0.9868          & 0.9827          & 0.9720          & 0.9452          & 0.9617          \\
LEMLL  & 0.9705          & 0.9451          & 0.9671          & 0.9491          & 0.9482          & 0.9274          & 0.9497          & 0.9467          & 0.9565          & 0.9619          & 0.9756          & 0.9723 \\ \hline\hline
\end{tabular}
\caption{  The recovery results on the real-world datasets measured by (Chebyshev$\downarrow$, Clark$\downarrow$, Canberra$\downarrow$, K-L $\downarrow$, Cosine$\uparrow$).}
\label{recover_main}
\end{table*}

\textbf{Comparing Methods}:  We compare LCdr against five  LE  approaches with parameter configurations suggested in respective literatures.
\begin{itemize}
    \item GLLE\cite{tkde/XuLG21}: GllE recovers LD directly from the logical label using the graph Laplacian method, where the regularisation parameter $\lambda$is set to 0.01  and the number of neighbors K is set to q+ 1.
    \item LESC\cite{Tang2020LabelEW}: LESC used a low-rank representation to capture the global relationship of the sample and predict the LD of the sample.  The parameters $\lambda_1$ and $\lambda_2$ are selected among $\{0.0001,0.001, \ldots, 10\}$.
    \item LEMLL\cite{DBLP:conf/icdm/ShaoXG18}: LEMLL L used the topology of the feature space to recover LD. We set the irrelevant marker to -1, the number of neighbors K is 3, instances whose distance computed is more than $\epsilon$ should be penalized as 10, and the type of kernel function is Lin.
    \item LP\cite{DBLP:journals/tkde/ZhangZFLG21}:LP used label propagation techniques to generate LD. We also choose the parameter $\alpha$ in LP to be 0.01.
    \item ML2\cite{DBLP:conf/aaai/HouGZ16}: ML2  Used the manifold technique to generated LD.  We set the number of neighbors K is set to q+ 1.
    \item LCdr: We set the number of neighbors K=10,  the threshold parameter $d^{\prime}$=10,  the trade-off parameter $\alpha$=$\beta$=0.1. 
\end{itemize}

\begin{table}[!h]\tiny\centering
\begin{tabular}{ccccccc}
\hline\hline
Metric       & LCdr          & LP   & ML2  & GLLE & LESC & LEMLL \\ \hline\hline
chebyshev    & \textbf{1.00} & 3.92 & 6.00 & 2.25 & 3.08 & 4.75  \\
clark        & \textbf{1.00} & 4.92 & 6.00 & 2.33 & 3.00 & 3.75  \\
canberra     & \textbf{1.00} & 4.83 & 6.00 & 2.33 & 2.83 & 4.00  \\
K-L        & \textbf{1.00} & 4.67 & 6.00 & 2.25 & 3.17 & 3.92  \\
cosine       & \textbf{1.17} & 4.25 & 6.00 & 2.33 & 3.33 & 3.92  \\
 \hline\hline
\end{tabular}
\caption{The average ranks of six algorithms on five measures in the LD recovered experiment.}
\label{ranktable}
\end{table}

\subsection{Results}
\subsubsection{LD Recovery Performance }
For quantitative analysis, table \ref{recover_main} tabulates the results of the five LE algorithms on all real-world datasets, and the best performance is highlighted.  Note that since each LE algorithm only runs once, there is no record of standard deviation.  The average ranks of six algorithms on five measures as shown in Table \ref{ranktable}.  To analyze whether there are statistical performance gaps among comparing algorithms, \emph{Wilcoxon signed-ranks test}\cite{jmlr/Demsar06}, which is a widely-accepted statistical test for comparisons of two algorithms over several datasets, is employed. Table \ref{wilcoxontest} summarizes the statistical test results and the p-values for the corresponding tests are also shown in the brackets. From the above results,
we have the following observations:
\begin{itemize}
    \item Compared to the five algorithms LP, ML2, GLLE, LESC, and LEMLL, the proposed method achieves better performance in all cases (12 datasets and five metrics).
    \item  The proposed approach achieves statistically superior performance against comparing methods in terms of the five metrics.
\end{itemize}

\begin{table*}[t!]\tiny\centering
\setlength{\tabcolsep}{3mm}
\begin{tabular}{ccccccccccc}
\hline\hline
\multicolumn{11}{c}{chebyshev $\downarrow$} \\ \hline
 Dataset      & GLLE            & LCdr-GLLE                            & ML2    & LCdr-ML2                             & LP     & LCdr-LP                              & LESC   & LCdr-LESC                            & LEMLL  & LCdr-LEMLL      \\ \hline
\multicolumn{1}{c|}{spoem}   & 0.1318          & \multicolumn{1}{c|}{\textbf{0.1261}} & 0.6195   & \multicolumn{1}{c|}{\textbf{0.1366}} & 0.2062  & \multicolumn{1}{c|}{\textbf{0.0828}} & 0.1364    & \multicolumn{1}{c|}{\textbf{0.1281}} & 0.1961     & \textbf{0.0899} \\
\multicolumn{1}{c|}{alpha}   & 0.3261          & \multicolumn{1}{c|}{\textbf{0.2375}} & 3.4836   & \multicolumn{1}{c|}{\textbf{0.2187}} & 0.8226  & \multicolumn{1}{c|}{\textbf{0.2116}} & 0.3085    & \multicolumn{1}{c|}{\textbf{0.2329}} & 0.7693     & \textbf{0.2129} \\
\multicolumn{1}{c|}{spo5}    & 0.1938          & \multicolumn{1}{c|}{\textbf{0.1848}} & 0.7556   & \multicolumn{1}{c|}{\textbf{0.2283}} & 0.2737  & \multicolumn{1}{c|}{\textbf{0.1446}} & 0.2022    & \multicolumn{1}{c|}{\textbf{0.1912}} & 0.2441     & \textbf{0.1458} \\
\multicolumn{1}{c|}{cdc}     & 0.3058          & \multicolumn{1}{c|}{\textbf{0.2751}} & 0.6439   & \multicolumn{1}{c|}{\textbf{0.5055}} & 0.4693  & \multicolumn{1}{c|}{\textbf{0.2411}} & 0.4154    & \multicolumn{1}{c|}{\textbf{0.2716}} & 0.3017     & \textbf{0.2551} \\
\multicolumn{1}{c|}{cold}    & 0.3071          & \multicolumn{1}{c|}{\textbf{0.3000}} & 1.4296   & \multicolumn{1}{c|}{\textbf{0.6960}} & 0.3881  & \multicolumn{1}{c|}{\textbf{0.2718}} & 0.3174    & \multicolumn{1}{c|}{\textbf{0.2586}} & 0.3058     & \textbf{0.2757} \\
\multicolumn{1}{c|}{diau}    & 0.3003          & \multicolumn{1}{c|}{\textbf{0.2974}} & 2.9377   & \multicolumn{1}{c|}{\textbf{0.2191}} & 0.7369  & \multicolumn{1}{c|}{\textbf{0.2130}} & 0.3020    & \multicolumn{1}{c|}{\textbf{0.2559}} & 0.6702     & \textbf{0.2179} \\
\multicolumn{1}{c|}{dtt}     & 0.1704          & \multicolumn{1}{c|}{\textbf{0.1434}} & 1.0435   & \multicolumn{1}{c|}{\textbf{0.1466}} & 0.3524  & \multicolumn{1}{c|}{\textbf{0.1096}} & 0.1700    & \multicolumn{1}{c|}{\textbf{0.1593}} & 0.3506     & \textbf{0.1349} \\
\multicolumn{1}{c|}{elu}     & 0.2916          & \multicolumn{1}{c|}{\textbf{0.2214}} & 1.6434   & \multicolumn{1}{c|}{\textbf{0.2225}} & 0.4669  & \multicolumn{1}{c|}{\textbf{0.2046}} & 0.2934    & \multicolumn{1}{c|}{\textbf{0.2321}} & 0.5848     & \textbf{0.2138} \\
\multicolumn{1}{c|}{heat}    & 0.1380          & \multicolumn{1}{c|}{\textbf{0.1380}} & 1.0491   & \multicolumn{1}{c|}{\textbf{0.1016}} & 0.3799  & \multicolumn{1}{c|}{\textbf{0.0727}} & 0.1407    & \multicolumn{1}{c|}{\textbf{0.1320}} & 0.3479     & \textbf{0.0840} \\
\multicolumn{1}{c|}{spo}     & 0.2839          & \multicolumn{1}{c|}{\textbf{0.2180}} & 2.7752   & \multicolumn{1}{c|}{\textbf{0.2106}} & 0.7119  & \multicolumn{1}{c|}{\textbf{0.1977}} & 0.2938    & \multicolumn{1}{c|}{\textbf{0.2403}} & 0.6604     & \textbf{0.2016} \\
\multicolumn{1}{c|}{flickr}  & 0.2064          & \multicolumn{1}{c|}{\textbf{0.1898}} & 1.3994   & \multicolumn{1}{c|}{\textbf{0.1871}} & 0.4321  & \multicolumn{1}{c|}{\textbf{0.1640}} & 0.2179    & \multicolumn{1}{c|}{\textbf{0.1975}} & 0.3964     & \textbf{0.1782} \\
\multicolumn{1}{c|}{twitter} & 0.2585 & \multicolumn{1}{c|}{\textbf{0.2581}} & 1.3311   & \multicolumn{1}{c|}{\textbf{0.2559}} & 0.3971  & \multicolumn{1}{c|}{\textbf{0.2313}} & 0.2744    & \multicolumn{1}{c|}{\textbf{0.2705}} & 0.3724     & \textbf{0.2471} \\ \hline
\multicolumn{11}{c}{ K-L $\downarrow$ } \\ \hline
\multicolumn{1}{c}{Datasets}       & \multicolumn{1}{c}{GLLE} & \multicolumn{1}{c}{LCdr-GLLE}        & \multicolumn{1}{c}{ML2} & \multicolumn{1}{c}{LCdr-ML2}         & \multicolumn{1}{c}{LP} & \multicolumn{1}{c}{LCdr-LP}          & \multicolumn{1}{c}{LESC} & \multicolumn{1}{c}{LCdr-LESC}        & \multicolumn{1}{c}{LEMLL} & \multicolumn{1}{c}{LCdr-LEMLL} \\ \hline
\multicolumn{1}{l|}{spoem}   & 0.0276                   & \multicolumn{1}{l|}{\textbf{0.0263}} & 0.2410                  & \multicolumn{1}{l|}{\textbf{0.0371}} & 0.0449                 & \multicolumn{1}{l|}{\textbf{0.0174}} & 0.0298                   & \multicolumn{1}{l|}{\textbf{0.0274}} & 0.0447                    & \textbf{0.0184}                \\
\multicolumn{1}{l|}{alpha}   & 0.0123                   & \multicolumn{1}{l|}{\textbf{0.0069}} & 1.8433                  & \multicolumn{1}{l|}{\textbf{0.0064}} & 0.0668                 & \multicolumn{1}{l|}{\textbf{0.0062}} & 0.0117                   & \multicolumn{1}{l|}{\textbf{0.0067}} & 0.0629                    & \textbf{0.0062}                \\
\multicolumn{1}{l|}{spo5}    & 0.0327                   & \multicolumn{1}{l|}{\textbf{0.0313}} & 0.2447                  & \multicolumn{1}{l|}{\textbf{0.0519}} & 0.0426                 & \multicolumn{1}{l|}{\textbf{0.0228}} & 0.0364                   & \multicolumn{1}{l|}{\textbf{0.0332}} & 0.0407                    & \textbf{0.0196}                \\
\multicolumn{1}{l|}{cdc}     & 0.0281                   & \multicolumn{1}{l|}{\textbf{0.0232}} & 0.0895                  & \multicolumn{1}{l|}{\textbf{0.0696}} & 0.0530                 & \multicolumn{1}{l|}{\textbf{0.0180}} & 0.0523                   & \multicolumn{1}{l|}{\textbf{0.0224}} & 0.0234                    & \textbf{0.0200}                \\
\multicolumn{1}{l|}{cold}    & 0.0301                   & \multicolumn{1}{l|}{\textbf{0.0271}} & 0.4202                  & \multicolumn{1}{l|}{\textbf{0.1313}} & 0.0436                 & \multicolumn{1}{l|}{\textbf{0.0238}} & 0.0318                   & \multicolumn{1}{l|}{\textbf{0.0216}} & 0.0281                    & \textbf{0.0217}                \\
\multicolumn{1}{l|}{diau}    & 0.0130                   & \multicolumn{1}{l|}{\textbf{0.0128}} & 1.4508                  & \multicolumn{1}{l|}{\textbf{0.0082}} & 0.0645                 & \multicolumn{1}{l|}{\textbf{0.0079}} & 0.0137                   & \multicolumn{1}{l|}{\textbf{0.0100}} & 0.0582                    & \textbf{0.0081}                \\
\multicolumn{1}{l|}{dtt}     & 0.0177                   & \multicolumn{1}{l|}{\textbf{0.0132}} & 0.3896                  & \multicolumn{1}{l|}{\textbf{0.0161}} & 0.0566                 & \multicolumn{1}{l|}{\textbf{0.0109}} & 0.0180                   & \multicolumn{1}{l|}{\textbf{0.0159}} & 0.0636                    & \textbf{0.0144}                \\
\multicolumn{1}{l|}{elu}     & 0.0259                   & \multicolumn{1}{l|}{\textbf{0.0166}} & 0.7066                  & \multicolumn{1}{l|}{\textbf{0.0182}} & 0.0533                 & \multicolumn{1}{l|}{\textbf{0.0160}} & 0.0283                   & \multicolumn{1}{l|}{\textbf{0.0183}} & 0.0900                    & \textbf{0.0170}                \\
\multicolumn{1}{l|}{heat}    & 0.0119                   & \multicolumn{1}{l|}{\textbf{0.0118}} & 0.4006                  & \multicolumn{1}{l|}{\textbf{0.0091}} & 0.0651                 & \multicolumn{1}{l|}{\textbf{0.0062}} & 0.0128                   & \multicolumn{1}{l|}{\textbf{0.0115}} & 0.0614                    & \textbf{0.0071}                \\
\multicolumn{1}{l|}{spo}     & 0.0124                   & \multicolumn{1}{l|}{\textbf{0.0075}} & 1.3624                  & \multicolumn{1}{l|}{\textbf{0.0077}} & 0.0640                 & \multicolumn{1}{l|}{\textbf{0.0070}} & 0.0136                   & \multicolumn{1}{l|}{\textbf{0.0091}} & 0.0610                    & \textbf{0.0071}                \\
\multicolumn{1}{l|}{flickr}  & 0.0164                   & \multicolumn{1}{l|}{\textbf{0.0141}} & 0.5468                  & \multicolumn{1}{l|}{\textbf{0.0153}} & 0.0558                 & \multicolumn{1}{l|}{\textbf{0.0126}} & 0.0184                   & \multicolumn{1}{l|}{\textbf{0.0151}} & 0.0519                    & \textbf{0.0141}                \\
\multicolumn{1}{l|}{twitter} & \textbf{0.0274}          & \multicolumn{1}{l|}{0.0278} & 0.4789                  & \multicolumn{1}{l|}{\textbf{0.0293}} & 0.0482                 & \multicolumn{1}{l|}{\textbf{0.0251}} & 0.0307                   & \multicolumn{1}{l|}{\textbf{0.0302}} & 0.0466                    & \textbf{0.0277}                \\ \hline\hline
\end{tabular}
\caption{The recovery experiments results  measured by chebyshev$\downarrow$ and  K-L$\downarrow$. For the LE algorithm $\mathcal{F} \in\{$ GLLE,LESC, ML2, LP, LEMLL $\}$, the performance of the $\mathcal{F}$-LCdr is compared against that of $\mathcal{F}$.  }
\label{aug_for_otherLE}
\end{table*}

\begin{table*}[!h]\tiny
\setlength{\tabcolsep}{3.5mm}
\centering
\begin{tabular}{ccccccccccccc}
\hline\hline
      & spoem           & alpha           & spo5            & cdc             & cold            & diau            & dtt             & elu             & heat            & spo             & flickr          & twitter         \\ \hline
\textbf{LF+LC} & \textbf{0.0859} & \textbf{0.0136} & \textbf{0.0916} & \textbf{0.0163} & \textbf{0.0535} & \textbf{0.0413} & \textbf{0.0366} & \textbf{0.0165} & \textbf{0.0432} & \textbf{0.0596} & \textbf{0.0614} & \textbf{0.0730} \\
\textbf{X+LF}  & 0.0881          & 0.0137          & 0.0926          & 0.0165          & 0.0547          & 0.0421          & 0.0367          & 0.0166          & 0.0434          & 0.0605          & 0.0656          & 0.0782          \\
\textbf{LF+L}  & 0.0874          & 0.0164          & 0.0976          & 0.0197          & 0.0517          & 0.0678          & 0.0376          & 0.0192          & 0.0460          & 0.0601          & 0.0691          & 0.0834          \\
\textbf{X+L}  & 0.0886          & 0.0191          & 0.1338          & 0.0305          & 0.0834          & 0.0791          & 0.0590          & 0.0218          & 0.0593          & 0.0723          & 0.3270          & 0.0906          \\ \hline\hline
\end{tabular}
\caption{Ablation experiments for recovery experiments measured by Chebyshev $\downarrow$. ($\mathbf{LC}$ representatives the label-confidence, $\mathbf{LF}$ representatives the lower-dim features, $\mathbf{X}$ representatives the original features, $\mathbf{L}$ representatives the logical labels.)}
\label{ablish}
\end{table*}

\begin{table}[!h]\setlength{\tabcolsep}{0.5mm}\tiny
\centering
\begin{tabular}{cccccc}
\hline\hline
Method & Chebyshev$\downarrow$ & Clark$\downarrow$ & Canberra$\downarrow$ & K-L $\downarrow$ & Cosine$\uparrow$ \\ \hline
LP & \textbf{ win}[4.88e-04] & \textbf{ win}[4.88e-04] & \textbf{ win}[4.88e-04] & \textbf{ win}[4.88e-04] & \textbf{ win}[9.77e-04] \\
ML2 & \textbf{ win}[4.88e-04] & \textbf{ win}[4.88e-04] & \textbf{ win}[4.88e-04] & \textbf{ win}[4.88e-04] & \textbf{ win}[4.88e-04] \\
GLLE & \textbf{ win}[4.88e-04] & \textbf{ win}[4.88e-04] & \textbf{ win}[4.88e-04] & \textbf{ win}[4.88e-04] & \textbf{ win}[4.88e-04] \\
LESC & \textbf{ win}[4.88e-04] & \textbf{ win}[4.88e-04] & \textbf{ win}[4.88e-04] & \textbf{ win}[4.88e-04] & \textbf{ win}[4.88e-04] \\
LEMLL & \textbf{ win}[4.88e-04] & \textbf{ win}[4.88e-04] & \textbf{ win}[4.88e-04] & \textbf{ win}[4.88e-04] & \textbf{ win}[9.77e-04] \\ \hline\hline
\end{tabular}
\caption{ Wilcoxon signed-rank test for LCdr against other LE algorithms in terms of five evaluation metrics. Significance level $\alpha$=0.05.}
\label{wilcoxontest}
\end{table}

\subsubsection{Ablation Studies }
In this section, ablation studies are conducted on all the 12 real-world benchmark data sets. The training and test settings of LCdr’s variant models are exactly the same as those of LCdr.
 Table \ref{ablish} shows the detailed experimental results in terms of Chebyshev$\downarrow$.

\emph{Effectiveness of the  label confidence strategy}. We implement a variant model of LCdr and named it as LF+L, which train the classifier using low-dimensional features (LF) and a logical label in the loss function \ref{LEmodel}. Results reported
in Table \ref{ablish} verify that LC as a more compact label-confidence can train a better classifier than using logical labele.

\emph{Effectiveness of the  dimensionality reduction  strategy}. We implemented a variant of LCdr named X+LC, which uses label-confidence to train the LE model in loss function \ref{LEmodel}. The results of table \ref{ablish} demonstrate that the data with the redundant features removed can improve the generalizability of the proposed method.

\emph{Effectiveness of the label confidence and dimensionality reduction  strategy}. We
implement a plain version of LCdr namend X+L, which training the LE model with raw data. The results demonstrated by Table \ref{ablish} show that our proposed data augmentation method can directly improve the performance of the LE algorithm.

\subsubsection{Parameter Sensitivity Analyse}
Fig. \ref{para} gives an illustrative example of how the performance
of LCdr changes in terms of each evaluation metric when the
value of the  parameter $\alpha$, $\beta$, $d^{\prime}$ and $k$  in the overall objective function changes. As shown in Fig. \ref{para}, the trade-off parameter $\alpha$ which controls the strength of retention of linear correlation in the projection feature matrix does affect the performance of LCdr.  Next,  changing the threshold parameter $d^{\prime}$ controls the dimensionality of the projection feature matrix and also affects the performance of LCdr.  Then, we changed the number of neighbors $K$ which controls the strength of the instance correlation and can also have an impact on the experimental results. Finally, by changing the trade-off parameter$ \beta$, we can see that changes in the parameter $\beta$ also have an impact on LCdr performance.  However, the performance is still relatively stable
as the parameter value changes within a reasonable range, which serves as a desirable property in using the proposed approach. Similar results can be observed in other datasets.
\begin{table*}[!h]\tiny\centering
\setlength{\tabcolsep}{3mm}
\begin{tabular}{ccccccccccccc}
\hline\hline

\multicolumn{13}{c}{Chebyshev$\downarrow$}                                                                                                                                                                                    \\ \hline
      & spoem           & alpha           & spo5            & cdc             & cold            & diau            & dtt             & elu             & heat            & spo             & flickr          & twitter         \\\hline
ours & \textbf{0.0188} & \textbf{0.0166} & \textbf{0.0797} & \textbf{0.0172} & \textbf{0.0619} & 0.0559 & \textbf{0.0448} & 0.0315 & 0.0453 & \textbf{0.0390} & \textbf{0.0293} & \textbf{0.0275} \\
GLLE & 0.1329 & 0.0316 & 0.0883 & 0.0292 & 0.1025 & \textbf{0.0517} & 0.1253 & 0.0369 & \textbf{0.0450} & 0.1156 & 0.0577 & 0.0333 \\
ML2 & 0.1193 & 0.0581 & 0.0983 & 0.0429 & 0.0995 & 0.1668 & 0.0515 & 0.0400 & 0.0909 & 0.0801 & 0.0864 & 0.1680 \\
LESC & 0.0223 & 0.1911 & 0.5122 & 0.2646 & 0.3109 & 0.3476 & 0.2976 & 0.2689 & 0.3890 & 0.3325 & 0.3151 & 0.3275 \\
LEMLL & 0.1152 & 0.0370 & 0.3487 & 0.0455 & 0.1274 & 0.0638 & 0.1523 & 0.0364 & 0.1078 & 0.0774 & 0.0640 & 0.0500 \\
LP & 0.1166 & 0.0253 & 0.2943 & 0.0291 & 0.1000 & 0.0535 & 0.1261 & \textbf{0.0208} & 0.0741 & 0.0400 & 0.0544 & 0.0368 \\ \hline
\multicolumn{13}{c}{clark$\downarrow$}                                                                                                                                                                                        \\ \hline
      & spoem           & alpha           & spo5            & cdc             & cold            & diau            & dtt             & elu             & heat            & spo             & flickr          & twitter         \\\hline
ours & \textbf{0.0268} & \textbf{0.3171} & \textbf{0.1430} & \textbf{0.2733} & \textbf{0.1814} & \textbf{0.4037} & \textbf{0.1121} & 0.3006 & \textbf{0.2116} & 0.1742 & \textbf{0.1727} & \textbf{0.1850} \\
GLLE & 0.1927 & 0.3823 & 0.1709 & 0.3329 & 0.2667 & 0.4047 & 0.3639 & 0.3882 & 0.2393 & 0.4001 & 0.3265 & 0.2208 \\
ML2 & 0.1696 & 1.0412 & 0.2007 & 0.8009 & 0.2406 & 0.7660 & 0.1156 & 0.5823 & 0.3408 & 0.4152 & 0.4591 & 0.6802 \\
LESC & 0.0340 & 2.1670 & 1.0370 & 1.7841 & 1.1521 & 1.4452 & 1.2464 & 1.7879 & 1.4865 & 1.3577 & 1.3406 & 1.1334 \\
LEMLL & 0.1647 & 0.6845 & 0.6173 & 0.4849 & 0.3407 & 0.4850 & 0.4879 & 0.3685 & 0.4397 & 0.2871 & 0.3534 & 0.3395 \\
LP    & 0.1667          & 0.5272          & 0.5100          & 0.3388          & 0.2567          & 0.4139          & 0.3649          & \textbf{0.2340} & 0.3084          & \textbf{0.1621} & 0.2690          & 0.2601          \\ \hline
\multicolumn{13}{c}{canberra$\downarrow$}                                                                                                                                                                                     \\ \hline
      & spoem           & alpha           & spo5            & cdc             & cold            & diau            & dtt             & elu             & heat            & spo             & flickr          & twitter         \\\hline
ours & \textbf{0.0379} & \textbf{1.1013} & \textbf{0.2307} & 0.9017 & \textbf{0.3515} & \textbf{1.0434} & \textbf{0.1755} & 0.8588 & \textbf{0.4525} & 0.3911 & \textbf{0.4110} & \textbf{0.3522} \\
GLLE & 0.2703 & 1.2533 & 0.2549 & 1.1266 & 0.4296 & 0.8376 & 0.6456 & 1.0829 & 0.5294 & 0.7307 & 0.7754 & 0.4761 \\
ML2 & 0.2394 & 1.0412 & 0.3075 & \textbf{0.8009} & 0.3998 & 1.9498 & 0.2016 & 1.8894 & 0.6714 & 0.9625 & 1.0598 & 1.7782 \\
LESC & 0.0469 & 9.0556 & 1.5971 & 6.7970 & 2.0269 & 3.6092 & 2.3057 & 6.6016 & 3.5585 & 3.1819 & 3.2136 & 2.6630 \\
LEMLL & 0.2321 & 2.6466 & 1.0284 & 1.5731 & 0.5462 & 1.1686 & 0.9135 & 1.1180 & 1.0179 & 0.6241 & 0.8758 & 0.7060 \\
LP    & 0.2349          & 1.9948          & 0.8617          & 1.1139          & 0.4114          & 0.8613          & 0.6432          & \textbf{0.6828}          & 0.6601          & \textbf{0.3498} & 0.6557          & 0.5675          \\ \hline
\multicolumn{13}{c}{K-L $\downarrow$}                                                                                                                                                                                       \\ \hline
      & spoem           & alpha           & spo5            & cdc             & cold            & diau            & dtt             & elu             & heat            & spo             & flickr          & twitter         \\\hline
ours & \textbf{0.0007} & \textbf{0.0111} & \textbf{0.0146} & \textbf{0.0101} & \textbf{0.0167} & 0.0468 & \textbf{0.0064} & 0.0140 & \textbf{0.0150} & 0.0106 & \textbf{0.0077} & \textbf{0.0076} \\
GLLE & 0.0357 & 0.0164 & 0.0206 & 0.0157 & 0.0377 & \textbf{0.0327} & 0.0626 & 0.0237 & 0.0193 & 0.0509 & 0.0275 & 0.0110 \\
ML2 & 0.0286 & 0.1144 & 0.0339 & 0.0833 & 0.0317 & 0.1750 & 0.0069 & 0.0496 & 0.0396 & 0.0584 & 0.0534 & 0.1205 \\
LESC & 0.0011 & 0.5978 & 1.3109 & 0.5224 & 0.6514 & 0.6951 & 0.9139 & 0.5493 & 0.8036 & 0.5994 & 0.4895 & 0.3719 \\
LEMLL & 0.0267 & 0.0548 & 0.3027 & 0.0352 & 0.0618 & 0.0586 & 0.1156 & 0.0203 & 0.0668 & 0.0286 & 0.0323 & 0.0256 \\
LP    & 0.0273          & 0.0314          & 0.1989          & 0.0162          & 0.0346          & 0.0352          & 0.0630          & \textbf{0.0079}         & 0.0320          & \textbf{0.0087} & 0.0201          & 0.0155          \\ \hline\hline
\end{tabular}
\caption{ The predictive results of testing instances  measured by (Chebyshev$\downarrow$, Clark$\downarrow$, Canberra$\downarrow$, K-L $\downarrow$) on the real-world datasets, where IIS-LDL is adopted as the predictive model.}
\label{predectLDL}
\end{table*}
\subsubsection{Extend To Other LE Approach}
\begin{figure}[]
\centering
\includegraphics[width=0.4\textwidth]{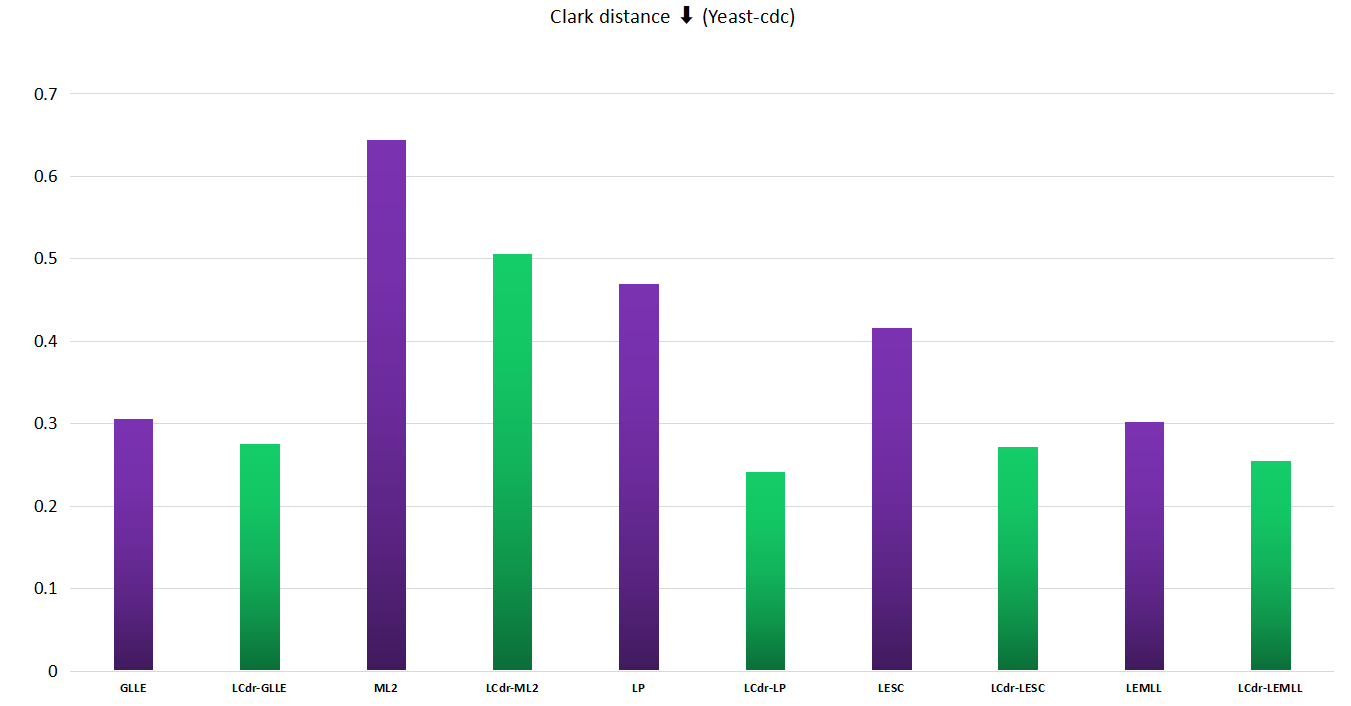} 
\caption{The recovery experiment  results visualized measured by Clark$\downarrow$ for each compared algorithm on the controlled Yeast-cdc dataset (with $\alpha$=0.1,$\eta$=10, k=10). For the LE algorithm $\mathcal{F} \in\{$ GLLE,LESC, ML2, LP, LEMLL $\}$, the performance of the $\mathcal{F}$-LCdr is compared against that of $\mathcal{F}.$}
\label{visualized}
\end{figure}

In the ablation experiments, we were surprised to find that the DA  strategy could directly improve the performance of the proposed method, and we made the conjecture of whether the DA strategy could be extended to other LE algorithms.  To test this conjecture, we compared the performance of different LE algorithms trained on LE data $\mathbf{\mathcal{T}(X, L)}$ and augmented data $\mathbf{\mathbbm{T}^{\prime}(\widetilde{X}, F)}$. The parameters are set as before. The results was shown in Table \ref{aug_for_otherLE}.  To observe the performance improvement more directly, we visualized the results for the control group, as shown in Fig. \ref{visualized}.  From the results, we can observe that:\begin{itemize}
    \item For all LE algorithms, the DA strategy can improve their performance in almost cases.
    \item In all cases, DA can improve the performance of these algorithms, which are ML2, LP, LESC, and LEMLL.
    \item DA can raise the performance of GLLE at 95.83$\%$ case.
\end{itemize}
The DA  strategy can improve the performance of most LE algorithms.  Based on this, better LC generation strategies and better dimensionality reduction methods may continue to improve the performance of LE in the future.

\subsubsection{LDL Predictive Results}
 In this section, we verify that the recovered LD can learn a good LDL model. The LDL predictive results were shown in Table \ref{predectLDL}. Here, IIS-LDL was used as the predictive model to generate the label distributions of testing instances for all the approaches, we chose the same parameter settings as the authors in the paper. From the experimental results, it can be seen that\begin{itemize}
     \item The proposed approach ranks 1st in 96.18$\%$ cases,  and it achieves superior performance against two algorithms including LESC and LEMLL in all cases.
     \item The proposed approach achieves superior performance against LP, GLLE, and ML2 in 85.42$\%$ , 93.75$\%$, and 97.92$\%$  cases, respectively.
     
 \end{itemize}

\section{Conclusion}
This paper  proposed a data augmentation approach to address two problems in LE
i.e. $(\mathbf{i})$: logical label do not provide more accurate supervision information and $(\mathbf{ii})$: feature redundancy in LE.  To be specific,   we used local consistency between instances to generate label-confidence and uses DR techniques to reduce the dimensionality of the feature space to remove redundant information.   Combining the above two, we proposed a non-linear LE model. Extensive experimental results demonstrate the effectiveness of the proposed approach consistently outperforms the other five comparing approaches.

\bibliographystyle{named}
\bibliography{ijcai22}

\end{document}